\documentclass[letterpaper, 10 pt, conference]{ieeeconf}  

\IEEEoverridecommandlockouts                              

\usepackage{graphics} 
\usepackage{epsfig} 
\usepackage{mathptmx} 
\usepackage{times} 
\usepackage{amsmath} 
\usepackage{amssymb}  
\usepackage{todonotes}
\usepackage{color}
\usepackage{romannum}
\usepackage{savesym} \savesymbol{AND}
\usepackage[noadjust]{cite}
\usepackage{hyperref}
\usepackage{algorithm, algorithmic}
 
\newcommand{\vth}{\mbox{\boldmath $\theta$}}

\newcommand{\vlambda}{\mbox{\boldmath $\lambda$}}

\newcommand{\vom}{\mbox{\boldmath $\omega$}}

\newcommand{\ve}{\mathbf e}
\newcommand{\vf}{\mathbf f}
\newcommand{\vg}{\mathbf g}
\newcommand{\vh}{\mathbf h}

\newcommand{\vl}{\mathbf l}

\newcommand{\vp}{\mathbf p}
\newcommand{\vq}{\mathbf q}
\newcommand{\vr}{\mathbf r}
\newcommand{\vs}{\mathbf s}

\newcommand{\vu}{\mathbf u}
\newcommand{\vv}{\mathbf v}

\newcommand{\vx}{\mathbf x}

\newcommand{\vA}{\mathbf A}
\newcommand{\vB}{\mathbf B}
\newcommand{\vC}{\mathbf C}
\newcommand{\vD}{\mathbf D}

\newcommand{\vF}{\mathbf F}

\newcommand{\vI}{\mathbf I}
\newcommand{\vJ}{\mathbf J}
\newcommand{\vK}{\mathbf K}
\newcommand{\vL}{\mathbf L}

\newcommand{\vP}{\mathbf P}
\newcommand{\vQ}{\mathbf Q}
\newcommand{\vR}{\mathbf R}
\newcommand{\vS}{\mathbf S}
\newcommand{\vT}{\mathbf T}

\title{\LARGE \bf
Real-Time Motion Planning of Legged Robots: A Model Predictive Control Approach 
}

\author{Farbod Farshidian, Edo Jelavi\'c, Asutosh Satapathy, Markus Giftthaler, Jonas Buchli  
\thanks{$^*$All authors are with the Agile \& Dexterous Robotics Lab, ETH Z\"urich, Switzerland, email: \{farbodf, jelavice, sasutosh, mgiftthaler, buchlij\}@ethz.ch}%
}

\begin{document}

\maketitle
\thispagestyle{empty}
\pagestyle{empty}

\begin{abstract}
We introduce a real-time, constrained, nonlinear Model Predictive Control for the motion planning of legged robots. The proposed approach uses a constrained optimal control algorithm known as SLQ. We improve the efficiency of this algorithm by introducing a multi-processing scheme for estimating value function in its backward pass. This pass has been often calculated as a single process. This parallel SLQ algorithm can optimize longer time horizons without proportional increase in its computation time. Thus, our MPC algorithm can generate optimized trajectories for the next few phases of the motion within only a few milliseconds. This outperforms the state of the art by at least one order of magnitude. The performance of the approach is validated on a quadruped robot for generating dynamic gaits such as trotting.
\end{abstract}

\section{Introduction}
One of the essential requirements for robust planning in real world applications is the capability of finding solutions in real-time to adjust the plan with the current state measurements. Many of today's online approaches have achieved this efficiency through task decomposition and model reduction approaches. The main idea behind these approaches is to decompose the locomotion problem into a set of simpler tasks which effectively reduces the number of control coordinates in each subtask. This simplification is the key to make the computation of the motion planner faster \cite{takanishi90, buschmann07} and facilitates finding solutions in real-time. Thus, these approaches are often used in most of the practical implementations of the Model Predictive Control (MPC) in legged robots \cite{wieber06, urata11, wittmann16, feng16, naveau17}. However, the simplification generally comes at the cost of limiting the maneuverability. This, in turn, can reduce the reachable set of solutions and renders the task synergy synthesis approaches overly conservative.

In contrast to the task decomposition approach, single task formulation offers the potential to treat the whole aspects of planning as a single problem without sacrificing performance. Due to the high complexity of legged systems, hand-designing the plan is often impractical. Therefore, an optimization method is often used to plan the robot's motion based on an user-defined performance index. This optimization problem is often formulated as an optimal control problem which provides the theoretical basis for designing a control policy. In general, there is no closed form solution for the optimal control problem with nonlinear system dynamics and cost function. Thus, the whole-body optimization often renders a computationally expensive numerical problem which can not be solved in real-time. 

While many of the applied optimization methods for motion planning of the legged robots do not scale favorably, the optimization methods based on using a Gauss-Newton Hessian approximation and a Riccati backward sweep demonstrate a great potential to be run in real-time on the high dimension problems. The notable examples of such approaches are DDP-based methods such as iLQR/G \cite{todorov05}, and SLQ (Sequential Linear Quadratic) \cite{sideris05}. Application of these algorithms in an MPC settings have been shown previously for a humanoid robot \cite{koenemann15}. In this work, we will show an implementation of a constrained SLQ algorithm in an MPC setting for motion planning of a quadrupedal robot. Unlike \cite{koenemann15}, we use a relatively longer time horizon due to our new approach which allows us to distribute the most expensive part of the computation in parallel.     

\subsection*{Contributions}
In this contribution, we introduce a constrained, nonlinear MPC approach for the motion planning of legged robots which its performance exceeding the current state of the art in robotics applications by at least one order of magnitude. Our MPC algorithm continuously re-optimizes the state and control input trajectories for the next few phases of the motion within only a few milliseconds. In order to achieve such a performance, we propose a variant of the SLQ algorithm which uses a multi-processing scheme for estimating value function in its backward pass. 

We demonstrate the performance of this algorithm for planning highly dynamic gaits such as trotting in an MPC fashion. The robustness and real-time planning capabilities of the approach is verified by inserting significant disturbances during execution. To the best of our knowledge, this work is the first to demonstrate a whole-body nonlinear MPC for periodic gait generation of the legged systems. Last but not least, our solver is available as an open-source software \cite{OCS2Library}. 

\section{Model Predictive Control Approach}
Our nonlinear MPC approach uses an efficient SLQ algorithm as its solver which can solve optimal control problems with nonlinear dynamics, cost, and equality constraints \cite{farshidian17a}. The performance of this approach for generating various gaits and obstacle avoidance has been shown previously in \cite{farshidian17a}. In this paper, we show how this algorithm can be used in an MPC loop. To this end, we first briefly introduce our optimization algorithm. Then, we propose a new approach which allows us to break the most computationally intensive part of the algorithm into several smaller calculations which can be carried out simultaneously. Finally, we discuss our MPC approach.        

\subsection{An Overview of the SLQ Algorithm}  
The continuous constrained SLQ algorithm is based on dynamic programming, which designs both a feedforward plan and a feedback controller through a quadratic approximation of the value function. Our continuous-time SLQ algorithm can handle state-input equality constraints through a Lagrangian method and state-only constraints through a penalty method. The complexity of the algorithm scales linearly with the time horizon of the optimization \cite{farshidian17a}. 

In general, planning algorithms based on Nonlinear Programming (NLP) require first to transcribe the infinite-dimensional, continuous problem to a finite-dimensional NLP. This discretization is often carried out using heuristics, which can result in numerically poor or practically infeasible solutions. In contrast, the continuous-time SLQ uses variable step-size ODE (Ordinary Differential Equation) solvers in its forward and backward passes. Given the desired accuracy, it can automatically discretize the problem using the error control mechanism of the variable step-size ODE solvers. Furthermore, using such an adaptive scheme -- in average -- decreases the number of discretized points in comparison to the discrete-time SLQ algorithm. This, in turn, can improve the run-time of an iteration since the number of calculations significantly decreases for the expensive operations such as linearization of the dynamics and the constraints.

In this paper, we use the SLQ algorithm for solving optimal control problem for switched systems. Switched systems are a subclass of a more general family known as hybrid systems. A hybrid system consists of a finite number of dynamical subsystems subjected to discrete events. These events are triggered either by an external input, or through intersection of the state trajectory to certain manifolds known as the switching surfaces. Upon triggering an event, a transition to a different subsystem takes place which can be followed by a sudden jump in the state vector. As more redistricted hybrid models, switched systems are characterized by continuous transitions of state trajectory during the switching moments. Here, we have further restricted our switched system model by assuming that the transitions are triggered by predefined switching times, in between predefined sequence of subsystems. For more general treatment of this problem refer to~\cite{farshidian17a, farshidian17b}. 

We formulate the constrained optimal control problem for switched systems in the finite time interval $[t_0,t_I]$ as
\begin{equation}
\label{eq:general_op}
\begin{aligned}
& \underset{\vu (\cdot)}{\text{minimize}}
& & \sum\limits_{i=0}^{I-1} {\Phi_i(\vx (t_{i+1})) + \int_{t_i}^{t_{i+1}} L_i(\vx, \vu, t) dt }  \\
& \text{subject to}
& & \dot{\vx} =  \vf_i(\vx, \vu), \hspace{5mm} \vx(s_0) = \vx_0, \hspace{2mm} \vx(t_i^-) = \vx(t_i^+) \\
& & & {\vg_1}_i(\vx, \vu, t) = 0, \  {\vg_2}_i(\vx, t) = 0,
\end{aligned}
\end{equation}
where $t_1$ to $t_{I-1}$ are the switching times and $I$ is the number of subsystems. For each mode $i$, the nonlinear cost function consists of a terminal cost, $\Phi_i(\cdot)$, and an intermediate cost, $L_i(\cdot)$. Here, $\vf_i(\cdot)$, ${\vg_1}_i(\cdot)$, and ${\vg_2}_i(\cdot)$ are respectively the system dynamics, the state-input constraints, and the state-only constraints in mode $i$.

The SLQ algorithm iteratively solves the extremal problem around the latest estimation of the optimal trajectories and improves the optimal control policy based on the solution of this local extremal problem. The local extremal problems are defined by the linearized system dynamics and constraints and the quadratic approximation of the cost function. The first step of each iteration is a forward integration of the system dynamics using the last approximation of the optimal controller. Next, a quadratic approximation of the cost function is calculated over the nominal state and input trajectories obtained from the forward integration. 
\begin{align}
& \widetilde{J} = \sum\limits_{i=0}^{I-1} {\widetilde{\Phi}_i(\delta\vx({t_{i+1}}))+ \int_{t_i}^{t_{i+1}} { \widetilde{L}_i(\delta\vx,\delta\vu,t)dt} } \notag \\
&\widetilde{\Phi}_i(\delta\vx) = \ q_{f,i} + \vq_{f,i}^\top \delta\vx + \frac{1}{2} \delta\vx^\top \vQ_{f,i} \delta\vx \notag \\
& \widetilde{L}_i(\vx,\vu,t) = q_i(t) + \vq_i(t)^\top \delta\vx + \vr_i(t)^\top \delta\vu  + \delta\vx^\top \vP_i(t) \delta\vu  \notag \\
& \hspace{17mm} + \frac{1}{2} \delta\vx^\top \vQ_i(t) \delta\vx + \frac{1}{2} \delta\vu^\top \vR_i(t) \delta\vu  
\label{eq:cost_quadratic_approximation}
\end{align} 
where $q_i(t)$, $\vq_i(t)$, $\vr_i(t)$, $\vP_i(t)$, $\vQ_i(t)$, and $\vR_i(t)$ are the coefficients of the Taylor expansion of the $i$th cost function in Equation~\eqref{eq:general_op} around the nominal trajectories. $\delta\vx$ and $\delta\vu$ are the deviations of state and input from the nominal trajectories. Constrained SLQ also uses linear approximations of the system dynamics and constraints in Equation~\eqref{eq:general_op} around the nominal trajectories.
\begin{align} 
& \delta\dot{\vx} = \vA_i(t)\delta\vx + \vB_i(t)\delta\vu \notag \\
& \vC_i(t)\delta\vx + \vD_i(t)\delta\vu + \ve_i(t) = \mathbf{0} \notag \\
& \vF_i(t)\delta\vx + \vh_i(t) = \mathbf{0} 
\label{eq:dynamics_linear_approximation}
\end{align}
Based on this LQ approximation, a generalized, constrained LQR algorithm is used to find an update to the feedback-feedforward control policy. For a more detailed discussion of the algorithm's derivation, refer to~\cite{farshidian17a}.

\begin{algorithm}[tpb] 
\caption{\textsc{FastSLQ} Algorithm} 
\label{alg:cslq}
\begin{algorithmic}[1] \scriptsize 
\STATE \textbf{Given:} 
\STATE \quad Initial stable control policy, $\{\vu_i(\vx,t)\}^{I-1}_{i=0} = \{\vu_{ff,i}(t) + \vK_i(t) \vx\}^{I-1}_{i=0}$
\STATE \quad Initial value function, $\{ V_i(\vx,t), V_{e,i}(\vx,t) \}^{I-1}_{i=0}$
\STATE \quad Heuristic function for approximating infinite time problem $V_I(\vx,t)=V^{lqr}(\vx)$
\REPEAT 
\STATE \textbf{Forward integrate the system dynamics} using adaptive step-size integrator.%
\STATE \qquad $\tau: \overline{\vx}(t_0),\overline{\vu}(t_0),\overline{\vx}(t_1),\overline{\vu}(t_1)\dots\overline{\vx}(t_{N-1}),\overline{\vu}(t_{N-1}),\overline{\vx}(t_N=t_{I})$
\FOR{$i = I-1$ to $0$ \textbf{in parallel}} 
\STATE \textbf{Quadratize cost function} along the trajectory $\tau$
\STATE \textbf{Linearize the system dynamics and constraints} along the trajectory $\tau$
\STATE Compute the constrained LQR problem coefficients
\STATE \qquad $\vD_i^\dagger = \vR_i^{-1} \vD_i^\top(\vD_i \vR_i^{-1} \vD_i^\top)^{-1}, \quad \widetilde\vA_i = \vA_i - \vB_i \vD_i^\dagger \vC_i $
\STATE \qquad $\widetilde\vC_i = \vD_i^\dagger \vC_i, \quad \widetilde \vD_i = \vD_i^\dagger \vD_i, \quad \widetilde \ve_i = \vD_i^\dagger \ve_i $
\STATE \qquad $\widetilde \vQ_i = \vQ_i + \widetilde\vC_i^\top \vR_i \; \widetilde\vC_i  - \vP_i \widetilde\vC_i - (\vP_i \widetilde\vC_i)^\top + \rho \vF_i^\top \vF_i $
\STATE \qquad $ \widetilde\vq_i = \vq_i - \widetilde \vC_i^\top \vr_i + \rho \vF_i^\top \vh_i, \quad \widetilde \vR_i = (\vI - \widetilde \vD_i)^\top \vR_i (\vI - \widetilde \vD_i) $
\STATE \qquad $\widetilde\vL_i = \vR_i^{-1} ( \vP_i^\top + \vB_i^\top \vS_i ) $
\STATE \qquad $\widetilde\vl_i = \vR_i^{-1} ( \vr_i + \vB_i^\top \vs_i ), \quad \widetilde \vl_{e,i} = \vR_i^{-1} \vB_i^\top \vs_{e,i} $
\STATE Calculate final value for Riccati-like equations
\STATE \qquad $\vS_i(t_{i+1})= \vQ_{f,i} + \frac{\partial^2}{\partial \vx ^2} V_{i+1}(\overline{\vx}(t_{i+1}),t_{i+1})$
\STATE \qquad $\vs_i(t_{i+1})= \vq_{f,i} + \frac{\partial}{\partial \vx}\! V_{i+1}\!(\overline{\vx}(t_{i+1}),t_{i+1})$,
$\vs_{e,i}(t_{i+1}) = \frac{\partial}{\partial \vx}\! V_{e,i+1}\! (\overline{\vx}(t_{i+1}),t_{i+1})$%
\STATE \qquad $s_i(t_{i+1})= q_{f,i} + V_{i+1}(\overline{\vx}(t_{i+1}),t_{i+1}) + V_{e,i+1}(\overline{\vx}(t_{i+1}),t_{i+1})$%
\STATE \textbf{Solve the final-value Riccati-like equations} in interval $[t_i, t_{i+1}]$
\STATE \qquad $- \dot \vS_i = \widetilde\vA_i^\top \vS_i + \vS_i^\top \widetilde\vA_i - \widetilde\vL_i^\top \widetilde\vR_i \; \widetilde\vL_i + \widetilde\vQ_i$
\STATE \qquad $- \dot \vs_i = \widetilde\vA_i^\top \vs_i - \widetilde\vL_i^\top \widetilde\vR_i \; \widetilde\vl_i + \widetilde\vq_i$
\STATE \qquad $- \dot \vs_{e_i} = \widetilde \vA_i ^\top \vs_{e_i} - \widetilde\vL_i^\top \widetilde\vR_i  \; \widetilde\vl_{e_i} + ( \widetilde\vC_i - \widetilde\vL_i )^\top \vR_i \; \widetilde\ve_i$
\STATE \qquad $- \dot s_i = q_i - \widetilde\vl_i^\top \widetilde\vR_i \; \widetilde\vl_i$
\STATE \textbf{Compute value function update}
\STATE \qquad $V_i(\vx, t) = s_i(t) + (\vx-\overline{\vx}(t))^\top \vs_i(t) + \frac{1}{2} (\vx-\overline{\vx}(t))^\top \vS_i(t) (\vx-\overline{\vx}(t))$%
\STATE \qquad $V_{e,i}(\vx, t) = (\vx-\overline{\vx}(t))^\top \vs_{e,i}(t)$%
\STATE \textbf{Compute controller update}
\STATE \qquad $ \vL_{i} = -(\vI - \widetilde\vD_{i}) \widetilde\vL_{i} - \widetilde\vC_{i}, \quad$
$ \vl_{i} = -(\vI - \widetilde\vD_{i}) \widetilde\vl_{i}, \quad$
$ \vl_{e,i} =  -(\vI - \widetilde\vD_{i}) \widetilde\vl_{e,i} - \widetilde\ve_{i}$
\STATE \qquad $ \vu_{ff,i} = \overline{\vu} + \alpha \vl_i + \vl_{e,i} - \vL_i \overline{\vx}, \quad \vu_i(\vx,t) = \vu_{ff,i}(t) + \vL_i(t) \vx$
\ENDFOR
\STATE \textbf{line search scheme}: optimize the learning rate, $\alpha$.
\UNTIL{convergence or maximum number of iterations}
\RETURN Optimized control policy and value function, $\{\vu_i(\vx,t), V_i(\vx, t), V_{e,i}(\vx, t)\}$
\end{algorithmic} 
\end{algorithm}

\subsection{\textsc{FastSLQ} Algorithm} \label{sec:disjointed_slq}
Each iteration of the SLQ algorithm consists of three main steps namely forward integration of system dynamics (i.e. forward pass), constructing the LQ approximation of the nonlinear problem, and solving Riccati-like equations (i.e. backward pass). Computing the LQ approximation in parallel has been proven to be essential in high dimensional problems. However, due to the sequential nature of integration, the forward and backward passes have been often implemented as single processes. In this section, we propose a variant of the SLQ algorithm which uses a parallel-processing scheme for calculating the backward pass. In practice, the backward pass is the most expensive part of the computations. Thus, parallel computation of the backward pass can significantly improve the speed of the algorithm. To this end, we refer to this algorithm as \textsc{FastSLQ}. 

In \textsc{FastSLQ}, we need first to divide the optimization time horizon into several disjoint intervals. A natural partitioning scheme for our switched system formulation is based on the switching moments but in general, any other partitioning approach can be considered. In order to solve the Riccati-like equations in each of these partitions, we should estimate the final value of the equations in each partition. 

A naive approach to compute the backward passes of these partitions in parallel is the  following. In each iteration, all processes -- in parallel -- integrate the Riccati-like equations of each partition backward in time. The final-value of these equations can be calculated based on the solution of the following partition in the previous iteration. This is in contrast to the sequential approach which waits until the computation of the following partition to complete. However, since the nominal trajectories of the previous iteration are different from the current ones, the Riccati-like equations in the subsequent iterations are different.  

To tackle this issue, we should first notice that SLQ uses the Bellman equation of optimality to locally estimate the value function. This local estimation is based on a quadratic approximation of the value function around the nominal trajectories. The coefficients of this quadratic model are calculated through the Riccati-like equations in the backward pass. \textsc{FastSLQ} leverages this approximation in order to correct the final values obtained from the previous iteration. To do so, it employs the value function approximation of the following partition from the previous iteration to estimate its value and its first order derivative at the partition's final time and the corresponding nominal state. Then, it uses the following equations to improve the estimation of the final values of the Riccati-like equations. 
\begin{align*}
& \vS^k_i(t_{i+1}) = \vQ^k_{f,i} + \tfrac{\partial^2}{\partial \vx ^2}  V^{k-1}_{i+1}(\overline{\vx}^k(t_{i+1}),t_{i+1}) \\
& \vs^k_i(t_{i+1}) = \vq^k_{f,i} + \tfrac{\partial}{\partial \vx}\! V^{k-1}_{i+1}\!(\overline{\vx}^k(t_{i+1}),t_{i+1}) \\
& \vs^k_{e,i}(t_{i+1}) = \tfrac{\partial}{\partial \vx}\! V^{k-1}_{e,i+1}\! (\overline{\vx}^k(t_{i+1}),t_{i+1}) \\
& s^k_i(t_{i+1}) = q^k_{f,i} + V^{k-1}_{i+1}(\overline{\vx}^k(t_{i+1}),t_{i+1}) + V^{k-1}_{e,i+1}(\overline{\vx}^k(t_{i+1}),t_{i+1})
\end{align*}
where we have defined 
\begin{align*}
& \delta\vx^{k}(\vx,t) = \vx-\overline{\vx}^{k}(t) \\
& V^{k}_{e,i}(\vx, t) = \delta\vx^{k}(\vx,t)^\top \vs^{k}_{e,i}(t) \\
& V^{k}_{i}(\vx,t) = s^{k}_i(t) + \delta\vx^{k}(\vx,t)^\top \vs^{k}_i(t) + \frac{1}{2} \delta\vx^{k}(\vx,t)^\top \vS^{k}_i(t) \delta\vx^{k}(\vx,t) ,
\end{align*}
the subscript $i$ and superscript $k$ refer to the subsystem index and the iteration number respectively. $t_i$s are the switching times (partitioning times), $V_i^k(\cdot)$ and $V_{e,i}^k(\cdot)$ are the value function approximation for the system in the null space and projected space of the constraints respectively. Finally, $\overline{\vx}^{k}(\cdot)$ is the nominal state trajectory in iteration $k$. 

\begin{figure}[t]
	\includegraphics[width=\columnwidth]{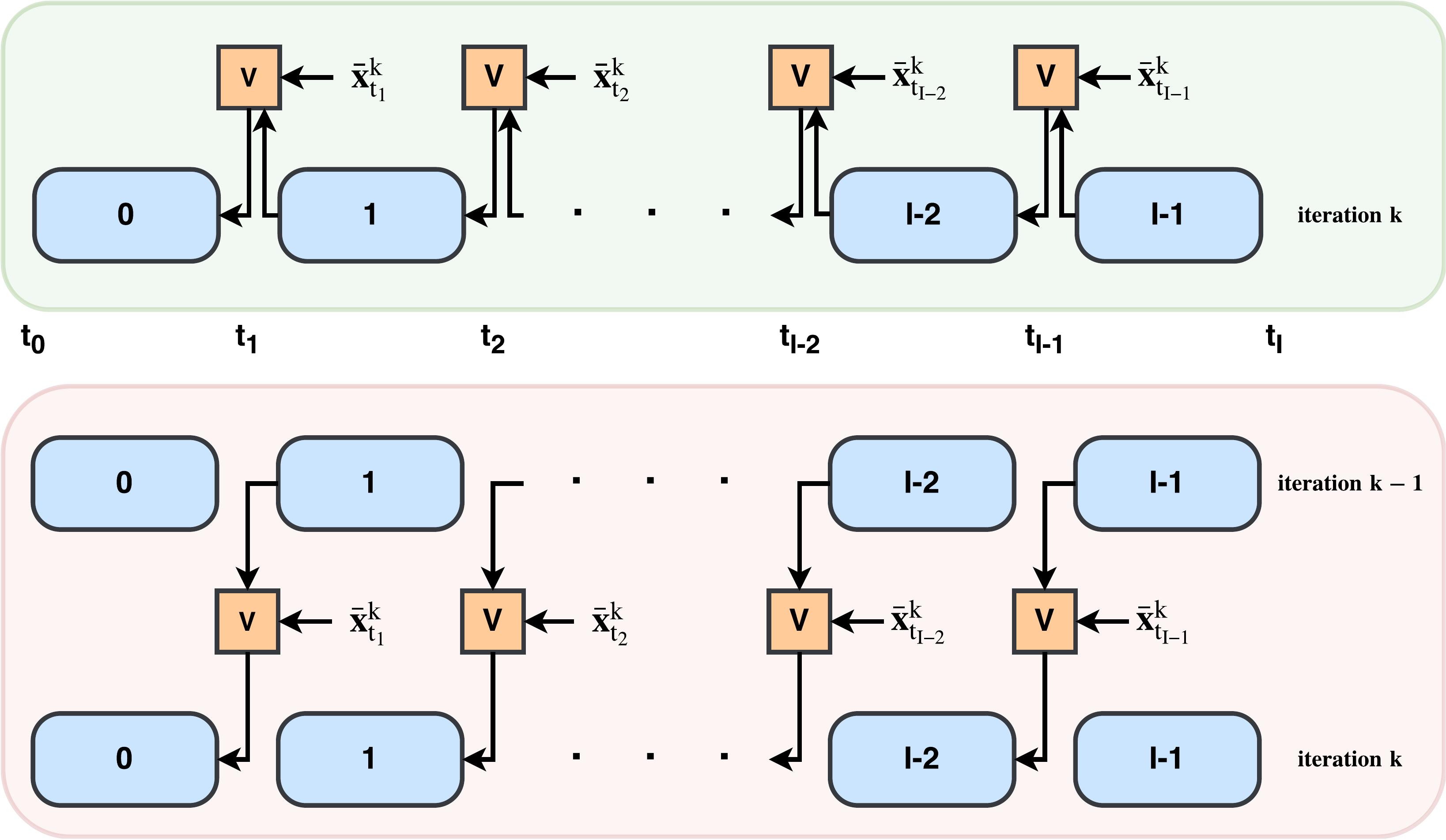}
	\vspace{-4mm}
	\caption{Comparison of the backward passes of SLQ and \textsc{FastSLQ} for estimating the value function based on solving Riccati-like equations. The top graph (green box) illustrates the sequential approach in which the Riccati-like equations are solved backward in time. The bottom graph (red box) illustrates the parallel computation approach. In this approach, instead of waiting for the solution of the neighboring partition, the approximation of the value function from the previous iteration is used. However, in order to account for the changes of the nominal trajectories in two iterations, the values function and its derivatives are re-evaluated at the current nominal trajectories.}
	\label{fig:ricatti}
\end{figure}

A high level illustration of the backward pass of \textsc{FastSLQ} is shown in Fig.~\ref{fig:ricatti}. Note that, since \textsc{FastSLQ} provides a quadratic approximation of the value function, the correction is only effective up to the first order terms (i.e. $\vs(\cdot),\vs_e(\cdot),s(\cdot)$) and the second order term (i.e. $\vS(\cdot)$) is directly used from the previous iteration. 

A summary of \textsc{FastSLQ}, is given in Algorithm~\ref{alg:cslq}. For a reliable implementation of \textsc{FastSLQ}, extra care should be taken when selecting the learning rate in the line search scheme. The line search scheme in the original SLQ algorithm favors the largest learning step which has a lower cost than the nominal cost. In order to ensure that the changes of the nominal trajectories in successive iterations are small enough that the quadratic approximation of the value function is valid, an extra criterion should be included in the line search scheme of the \textsc{FastSLQ} algorithm. To this end, the following criterion is appended to the line search scheme's conditions. The new cost associated with the updated control policy should be close enough to the expected cost which is estimated through the approximate value function. This line search scheme is in particular necessary in the initial iterations when the solution is far from the optimal one. \footnote{Note that this is not a major issue in a real-time iteration MPC since the optimal solution of the subsequent problems are in the close neighborhood.}



The backward pass of the SLQ-type algorithms can be considered as a bootstrapping approach for estimating the value function. In the original SLQ algorithm, the state sweeps are carried out in an especial order -- from the final state towards the initial state. Due to the specific structure of the problem, this process converges in a single sweep. In contrast, \textsc{FastSLQ} uses a different sweeping order which allows to implement the process in parallel. At a very high level, this is similar to how the value function is estimated in asynchronous dynamic programing approach where the state sweep can be carried out in an arbitrary order as long as all states are visited.



\subsection{Dealing with State-Input Inequality Constraints}  \label{sec:inequality_constraints}
In order to deal with state-input inequity constraints, we employ an approach similar to \cite{tassa14}. While this method does not necessarily result in an optimal solution, it ensures that the planned trajectories are constraint-satisfactory. In this method, if at any time during the algorithm's forward pass, an inequity constraint is violated, the control vector is projected on the plane defined by the linearized inequality constraint.  
   
\subsection{\textsc{FastSLQ-MPC} with Real-Time Iteration Scheme} \label{sec:fastslq-mpc}
The real-time scheme is proposed for the scenarios such as MPC where we need to solve a sequence of optimization problems, but we do not have the time to iterate each problem to convergence. In the real-time iteration MPC, the optimal control solver never attempts to iterate to convergence, but instead only takes one iteration towards the solution of the most current MPC problem (triggered with new measurements), before proceeding to the next one \cite{diehl05}. 

In the absence of disturbances and if the terminal time of the optimization is not receded, this scheme subsequently delivers approximations of the optimal policy that become better and better over iterations. However, in the presence of disturbances or receding horizon MPC scheme, it is crucial to the success of this method that the transition between subsequent problems be carefully designed.   

In the presence of disturbances, the initial state of the subsequent optimization problem would deviate from the planned state. Thus, in a naive real-time iteration implementation, depending on the amplitude of the deviations, the convergence can be drastically affected. In order to reduce this effect, the current solution of the optimizer should be corrected based on the state deviations. In the \textsc{FastSLQ-MPC} algorithm, such a correction can be achieved using its locally linear feedback policy. This policy generalizes the optimal open loop solution to a vicinity of the optimal trajectories while respecting constraints. Thus local adjustment of the plan can be realized by using the optimal feedback policy of the SLQ instead of using optimized trajectories. 

In addition to disturbance, we should also address the issue arises from the shifting of the optimization's terminal time in order to maintain the time horizon of optimization. This issue arises because of the finite-time optimization setup. To tackle it, we increment the terminal cost of the optimization with the value function of a fictitiously infinite-time LQR solution defined at the final state. Moreover, we have employed a varying time horizon scheme, in which the final time is set to a future switching time such that the time horizon includes exactly $n$ complete switching modes. For example, for $n=2$ at time $t$ where subsystem $i$ is active ($t_i \leq t < t_{i+1}$), the terminal time will be set to $t_{i+3}$. This means that if the average activation time of subsystems is $\Delta t$, the optimization time horizon varies in between $2\Delta t$ and $3 \Delta t$. 

In this way, the terminal time of the MPC optimization is always set to moments of mode switch. As long as the final states of the modes are controllable, they can be stabilized by fictitiously linear LQR controllers. Thus, these LQR problems have finite value functions. Defining such a stable LQR problem also allows using a quasi-infinite horizon MPC approach \cite{chen98, morari99}. The stability of this quasi-infinite horizon MPC can be guaranteed as long as the system is controllable at the terminal time. For the gait studied in this paper, this switching moments correspond to 4-leg stance mode in which the system is fully actuated and controllable. Our MPC approach is described in Algorithm~\ref{alg:mpc}.

\begin{algorithm}[t] 
\caption{\textsc{FastSLQ-MPC} Algorithm} 
\label{alg:mpc}
\begin{algorithmic}[1] \scriptsize 
\STATE \textbf{Given:} 
\STATE \quad Initial stable control policy, $\{\vu_i(\vx,t)\} = \{\vu_{ff,i}(t) + \vK_i(t) \vx\}$. 
\STATE \quad LQR Quadratic cost function, $\{\vQ^{lqr}_i,\vR^{lqr}_i \}$, for heuristic value function. 
\REPEAT
\STATE \textbf{Get the current time and state}, $t_0, \vx_0$.
\IF {$t_f - t < t_h$}
\STATE \textbf{Append a new subsystem}, $I-1$, based on the gait pattern.  
\STATE \textbf{Adjust final time}: $t_f = t_{I}$
\STATE \textbf{Calculate an LQR} for subsystem $I$ using linearized system at $\vx(t_{I-1})$.%
\STATE \textbf{Append control policy} with LQR controller: $\vu_{I-1}(\vx,t) = K^{lqr} (\vx - \vx(t_I) )$.%
\STATE \textbf{Set final cost} using LQR quadratic value function: $V_{I}(\vx,t) = V^{lqr}(\vx)$.%
\ENDIF
\STATE \textbf{Adjust the previous controller}
\STATE \quad Rollout from $(t,\vx)$ using policy $\{\vu_i(\vx,t)\}$ and get nominal trajectories
\STATE \qquad $\tau: \overline{\vx}(t_0),\overline{\vu}(t_0),\overline{\vx}(t_1),\overline{\vu}(t_1)\dots\overline{\vx}(t_{N-1}),\overline{\vu}(t_{N-1}),\overline{\vx}(t_N=t_{I})$
\STATE \quad Adjust the previous controller feedforward component, $\{\vu_{ff,i}(t)\}$.
\STATE \qquad $\vu_{ff,i}(t) = \overline{\vu}(t) - \vK_i(t) \overline{\vx}(t)$
\STATE \textbf{Perform single iteration of SLQ} 
\STATE \quad Set adjusted policy, $\{\vu_i(\vx,t)\}^{I-1}_{i=0} = \{\vu_{ff,i}(t) + \vK_i(t) \vx\}^{I-1}_{i=0}$. 
\STATE \quad Set augmented value function, $\{ V_i(\vx,t), V_{e,i}(\vx,t) \}^{I}_{i=0}$. 
\STATE \quad Perform single iteration of SLQ in time period $[t_0, t_f]$.
\STATE \textbf{Get the feedback policy and value funtion}
\STATE \quad $\{\vu_i(\vx,t)\}_{i=0}^{I-1}$, $\{ V_i(\vx,t), V_{e,i}(\vx,t) \}^{I}_{i=0}$.
\UNTIL {finished}
\end{algorithmic} 
\end{algorithm}

\section{Experimental Setup} 
\subsection{Platform Description}
For evaluating our MPC approach, we use a hydraulically actuated quadruped robot known as HyQ \cite{semini11}. HyQ is a fully torque controlled quadruped and it features three joints per leg. All joints are equipped with absolute and relative encoders. The joint torques are measured by load cells which are also used to estimate the contact forces.

In our setup, we use a dedicated computer to run the MPC control loop. This computer has an Intel Corei7 4790 processor (8~M Cache, 3.60~GHz processor frequency). The MPC loop receives the current state of the robot from the midlevel control computer that executes the tracking controller described in subsection~\ref{sec:mc}. This tracking controller runs at 250~Hz. The midlevel controller then sends desired torque to a lowlevel torque controller. In return, it receives current state measurements and computes the base and ground state estimates. The lowlevel controller runs a torque tracking control loop for actuators at 1~KHz. 

In order to estimate the base state, we use a Kalman filter approach introduced in \cite{bloesch13}. For estimating the ground plane, we use a simple approach which approximates the ground plane by passing a plane at the stance feet of the robot (in two consequential phases). 

\subsection{Modeling Framework} \label{sec:model}

In this paper, we choose to use a whole-body modeling approach known as the Center of Mass (CoM) dynamics plus full kinematics \cite{dai14}. This model includes 12 states describing the CoM motion as well as robot's joint positions which describe the full kinematic of the robot. The control input of this model includes the contact forces at end-effectors and the joint velocities. Due to the impact forces at the touch-down moments of swing legs, the state trajectories are not continuous. Thus, this model is a hybrid model. However, here, we model the legged robot as a switched system.     

This transition from a hybrid model to a switching model requires to assume that there is no state jump at the moments of phase transitions. Here, we reinforce such an assumption by designing swing leg trajectories with zero approaching velocity at the touch-downs. Assuming that the phase sequence of the motion is predefined, we can construct a switched model with a set of constraints on the velocities and the contact forces at end-effectors \cite{farshidian17a}. For HyQ, The CoM dynamics plus full kinematics model has 24 states and 24 inputs. It includes 12+\textsc{NumberOfSwingLegs} active state-input equality constraints at each phase of motion. The equation of motion for this model is as following 
\begin{align*}
&\left\{ 
\begin{array}{ll}
		\dot{\vth} = \vT(\vth) \left( \vom - _B\vJ_{com}^{\omega}(\vq) \; \vu \right) \\
		\dot{\vp}  = \vR(\vth) \, \vv \\
		\dot{\vom} = \vI^{-1}(\vq) \left( \dot\vI(\vq,\vu) -\vom \times \vI(\vq)\vom + \sum_{i=1}^4{ {\vr_f}_i(\vq) \times {\vlambda_f}_i} \right)\\
		\dot{\vv}  = \vg(\vth) + \frac{1}{m} \sum_{i=1}^4{\vlambda_f}_i \\
		\dot{\vq} = \vu 
\end{array}	 
\right. \notag \\
&\left\{ 
\begin{array}{ll}
		{\vu_f}_i(\vq,\vu) = \mathbf{0}   &\text{if $i$ is a stance leg} \\
		{\vu_f}_i(\vq,\vu) \cdot \hat{n} = c(t), \quad  {\vlambda_f}_i = \mathbf{0} \quad  &\text{if $i$ is a swing leg} \\
\end{array}	 
\right.
\end{align*}
where $\vT$ is a matrix which transforms the angular velocities in the base frame to the Euler angles derivatives in the global frame. $\vR$ is the rotation matrix of the base with respect the global frame, $\vg$ is the gravitational acceleration in the body frame, $\vI$ and $m$ are respectively the total moment of inertia about the CoM and the total mass. $_B\vJ_{com}^{\omega}$ is the Jacobian matrix of CoM rotation with respect to robot's base frame. $\vr_{f_i}$, $\vu_{f_i}$, $\vlambda_{f_i}$ are respectively the position, velocity and contact force vector of foot $i \in \{1,2,3,4\}$. 

We use a predefined swing leg trajectory, $c(t)$, in the orthogonal direction of the contact surface ($\hat{n}$) which ensures that the touch-down takes place according to the predefined switching times. Furthermore, it ensures that the velocity of the swing foot before the contact is zero. The unilateral and the friction constraints on contact forces are enforced by the method introduced in subsection~\ref{sec:inequality_constraints}.  

\subsection{Model Comparison}
The computational complexity of SLQ scales cubically with respect to the sum of state's and input's dimensions, $O\left((n_x+n_u)^3\right)$. In our switched model formulation for HyQ, this sum is of dimension $48$. A comparable model of HyQ which results in a same computational complicity is the rigid body modeling approach with a soft contact model \cite{neunert17}. The state space in this model is of dimension 36 which consists of the base pose and twist as well as joint angles and velocities. The control inputs only includes the joint torques with dimension 12. The contact forces are calculated using a soft model consisting of nonlinear springs and dampers. 

Each of these models has their advantages and disadvantages. The switched model uses hard constraints to satisfy contact unilateral constraints. However, the full model with soft contacts uses a penalty method in which constraints are not always fully satisfied. Thus, special care should be taken to implement them on hardware and in general achieving a good go-to task is hard. On the other hand, the optimized plan for the swing leg velocities in the full model with soft contact is always continuous but the velocities of the swing legs in the switched model are piecewise continuous. Therefore, in the switched model, we need to pre-filter the planned joint velocities before applying them on hardware.

\begin{figure} [tbp]
    \includegraphics[width=\columnwidth]{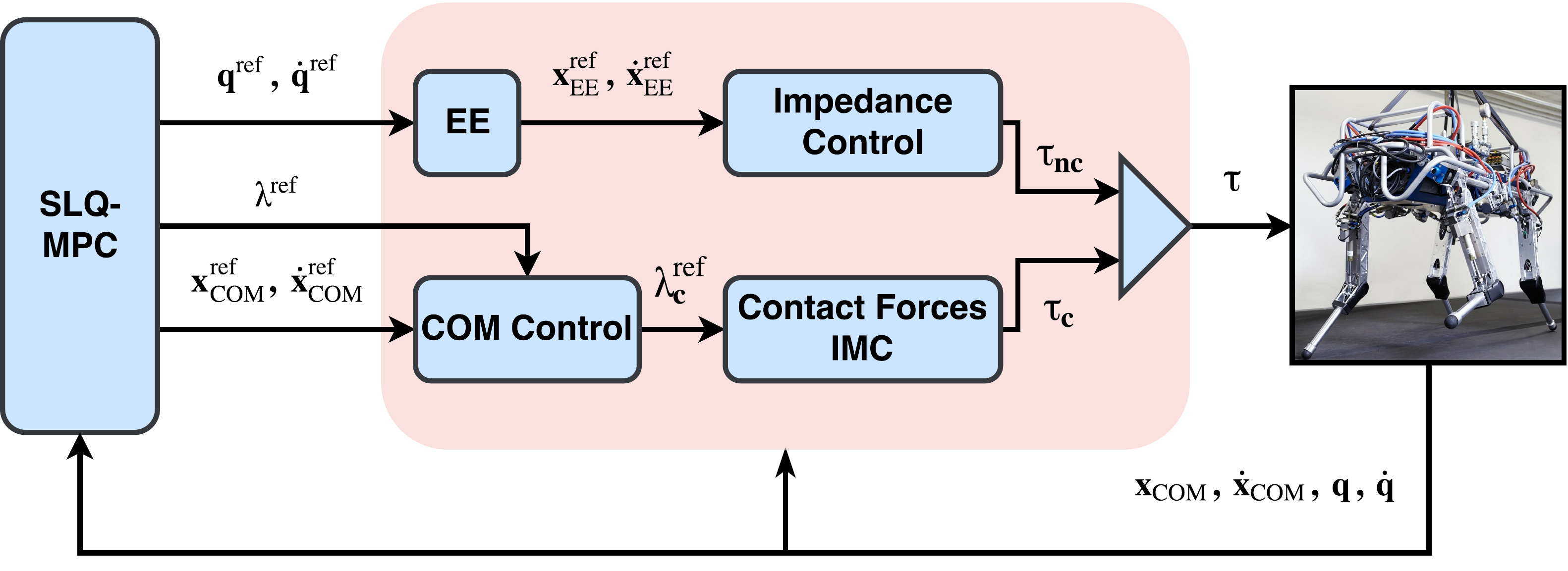}
    \vspace{-4mm}
    \caption{Overview of the motion control and planning structure. The planner is illustrated at far left and the motion controller (red block) in middle. The EE block transforms the joint's coordinate to the end-effector's Cartesian coordinate which is controlled by an impedance controller (inverse-dynamics + PD). The CoM controller stabilizes the CoM by providing correction to the CoM acceleration which is then mapped to an equivalent correction to $\vlambda_{ref}$ of the MPC. Finally, the IMC tracks the corrected end-effector's forces.   
    }
    \label{fig:mc_structure}
\end{figure}

\subsection{Motion Control Structure} \label{sec:mc}
In our switched model approach the contact forces at end-effectors are part of the control authority. However, the actual commands to the robot are the joint torques. There are two ways to realize this contact forces. Either, we should map them back to an equivalent joint torques or we should directly control the contact forces. Here, we have chosen the direct approach. To this end, we use a robust motion control approach introduced in \cite{farshidian17c} which relies on the robust tracking of contact forces in face of rigid body model mismatch, actuator dynamics, delays, contact surface stiffness, and unobserved ground profiles. 

Fig.~\ref{fig:mc_structure} demonstrates an overview of the motion controller introduced in \cite{farshidian17c}. In order to manipulate the contact force directly, this structure uses an especial system decomposition which allows to control the swing leg trajectories and contact forces independently. This motion control structure consists of three main components: contact force controller, swing leg controller, and CoM controller. The contact force controller uses a robust Internal Model Controller (IMC) to track the planned contact forces. The swing leg controller uses an inverse dynamics scheme plus a PD controller to track the desired end-effector motion in the Cartesian space. Finally, the CoM controller is responsible for tracking desired, feasible motions of CoM.

In theory, if the MPC loop runs fast enough (e.g. 250~Hz), its re-planing scheme should be able to stabilize the CoM. However, in practice our MPC loop run at 60~Hz which is slower than it can be used as a stabilizing controller. In this case a CoM tracking controller is employed in order to deal with the discrepancies between the model and hardware. Our CoM controller uses a PD feedback controller on the planned trajectories which provides correction to the CoM acceleration which is later mapped to an equivalent correction to the MPC planned contact forces.

\section{Results}
In this section, we show how \textsc{FastSLQ-MPC} can be applied for planning of the periodic gait patterns on real hardware. We also demonstrate the robustness and re-planning capabilities of the approach by adding significant disturbances during execution. Finally, we benchmark the run-time speed of our \textsc{FastSLQ} implementation.

\begin{figure}[t]
	\includegraphics[width=1\columnwidth]{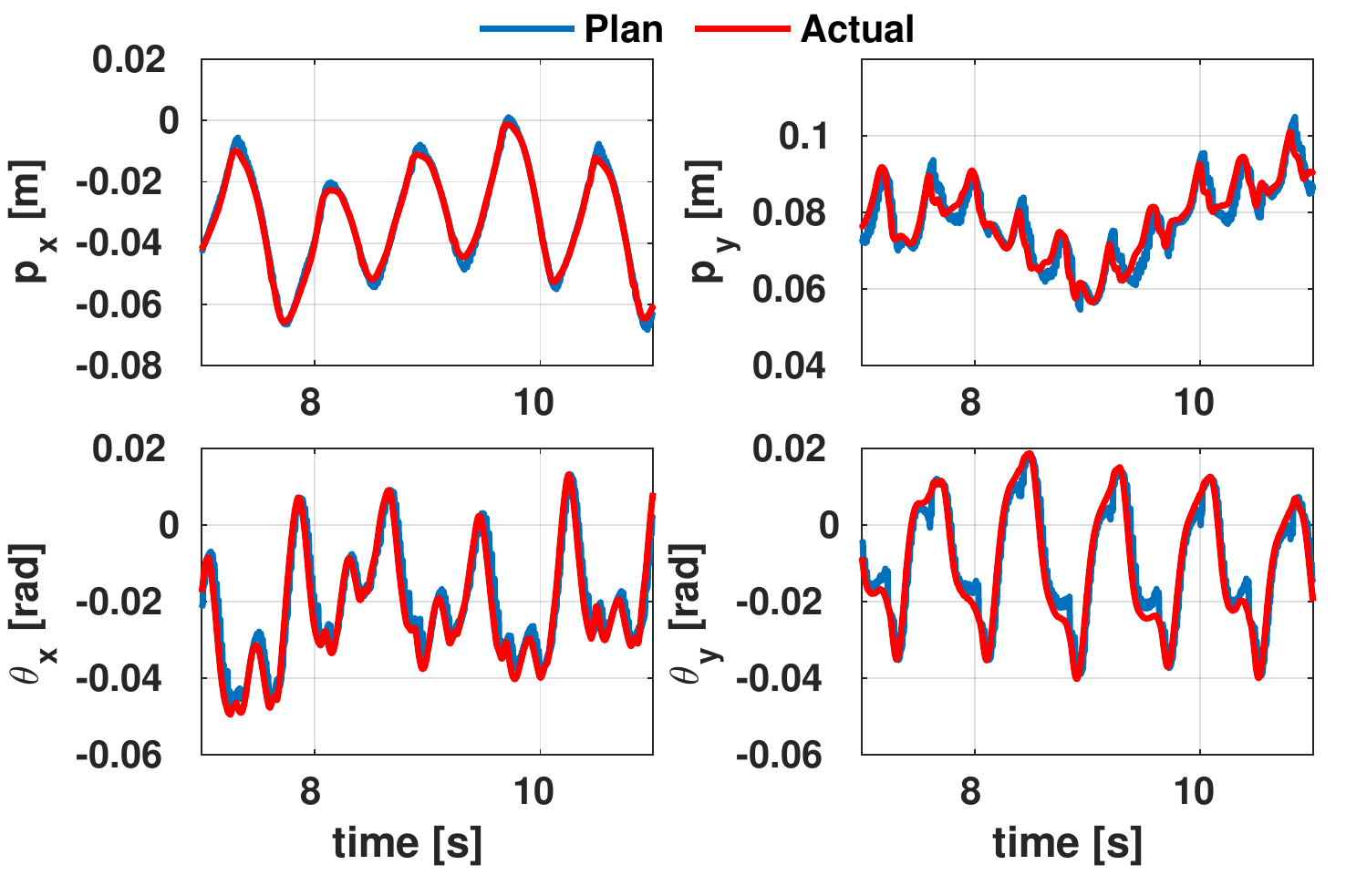}
	\vspace{-8mm}
	\caption{CoM pose plots for HyQ trotting in-place. The blue lines are the planned references and the red lines are the actual states. The robot nicely maintains its pose and the position and orientation angles are varying in a small range. The periodic pattern of motion is clearly visible in these plots. }
	\label{fig:com}
\end{figure}

\begin{figure}[t]
	\includegraphics[width=1\columnwidth]{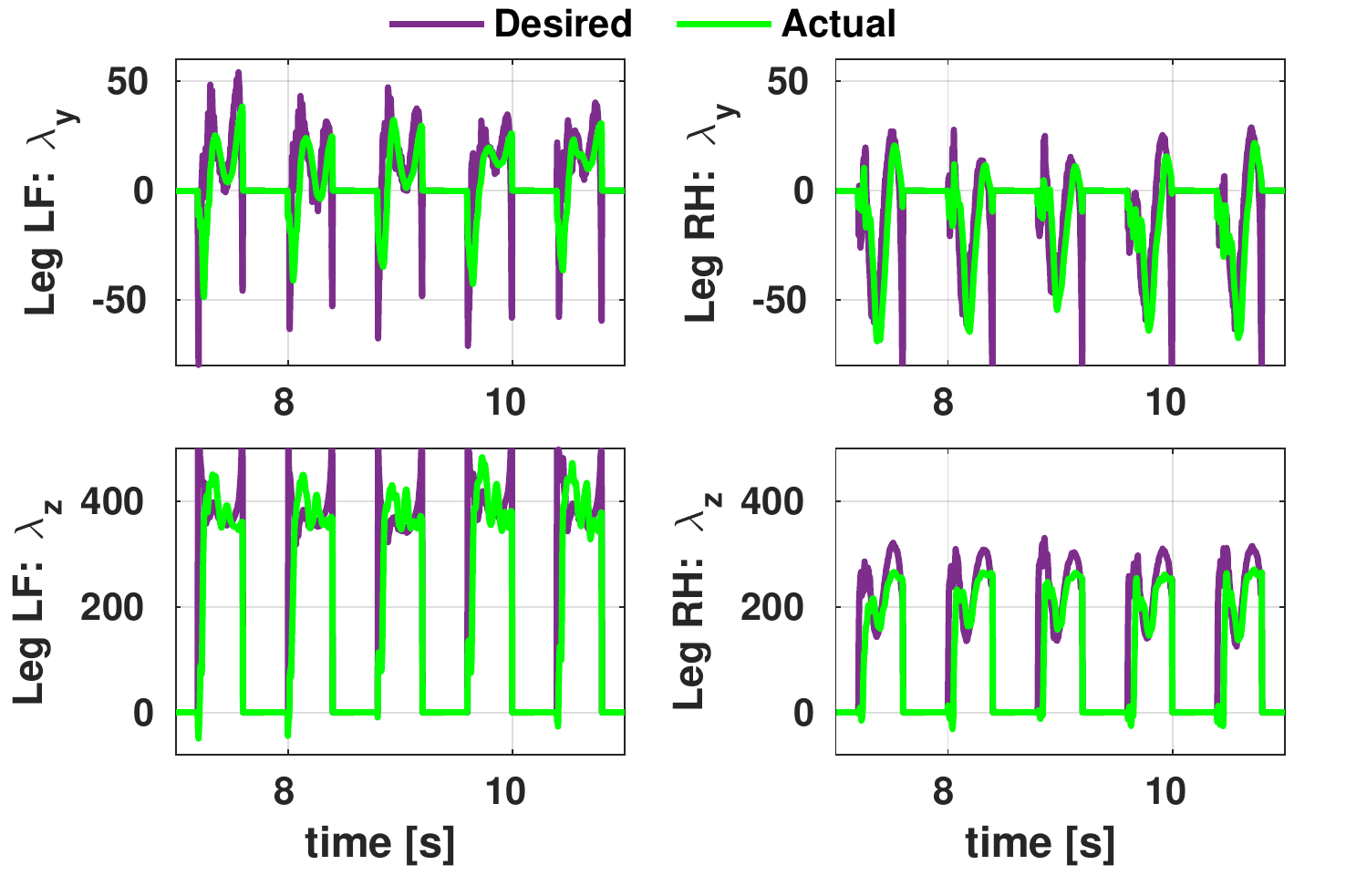}
	\vspace{-8mm}
	\caption{Contact forces for the trotting in-place task. The planned forces are in purple and the estimated forces are in green. The contact forces are relatively smooth during support phases, which facilitates hardware experiments. The contact forces also nicely reflect the stepping pattern.}
	\label{fig:contact}
\end{figure}

\subsection{Motion Planning for HyQ through \textsc{FastSLQ-MPC}}
In order to assess the performance of our MPC algorithm, we have performed a series of experiments on HyQ. All of the experiments presented in this section are implemented on hardware as well as two different physics engines namely SL and Gazebo. The performance of the robot was consistent across different simulators and was comparable to the hardware results. Due to the limited space, here, we only focus on the hardware results. However, in the attached video\footnote{\url{https://youtu.be/EYGVmcd9uds}}, more results are presented in both SL and Gazebo simulators. 

Three different experiments are presented in this section: ``trotting in-place task'', ``go-to task'', and ``disturbance rejection task''. Here, we mainly focus on the trotting gait. We also assume that the duration of each phase of motion is $400$ [ms] regardless of the given task. We use similar cost functions for all of the experiments. The cost function in each phase of motion (each subsystem) has a simple quadratic form. 
\begin{align*}
J_i =& \frac{1}{2} \left( \vx(t_{i+1})-\vx_d(t_{i+1}) \right)\!^\top\! \vQ_{f,i} \left( \vx(t_{i+1})-\vx_d(t_{i+1}) \right)  \\
&+ \int_{t_i}^{t_{i+1}} { \frac{1}{2} \vx(t)\!^\top\! \vQ_{i} \vx(t) + \frac{1}{2} \vu(t)\!^\top \vR_{i} \vu(t) \, dt}
\end{align*}
where $\vQ_{f,i}$, $\vQ_{i}$, and $\vR_{i}$ are constant, diagonal matrices of appropriate dimensions. $\vx_d(t_{i+1})$ defines the desired goal state which can be used by user to command the robot to move around. This variable is constant for the in-place trotting task and it is set to the commanded goal state for the go-to task. For all of the results presented in this section, we use 2 phases ahead setting for the MPC time horizon. In this case, the time horizon varies from $0.8$ [s] to $1.2$ [s] (refer to the discussion in \ref{sec:fastslq-mpc}). Using the four independent threads of our processor, the MPC loop runs at about 60~Hz. In the following, we discuss our results in details.     
 
\paragraph{Trotting in-place task}
In this task, HyQ trots in-place and tries to maintain its pose. Fig~\ref{fig:com} shows the xy position of CoM as well as the roll and the pitch angles of the base. The blue lines are the planned references and the red lines are the actual states. As you see, the references are nicely tracked and the position and orientation angles are varying in a small range. The periodic pattern of motion is clearly visible in these plots.

Fig~\ref{fig:contact} shows the yz-direction contact forces at the same time period for two opposing legs namely left-front, LF, leg and right-hind, RH, leg. The planned forces are in purple and the estimated forces are in green. The contact forces nicely visualize the stepping sequence where the discontinuities in the contact forces concur with a moment of touch-down or lift-off. Due to the imbalance of the robot's weight, the force profiles of the two legs have different patterns.   
 
\paragraph{Go-to task} 
In this task, after the robot starts trotting, we command it to move $1$~[m] ahead ($x$ direction). The time horizon in which the task should be completed is calculated based on a heuristic. For a motion with a predefined stepping time, the problem of estimating time horizon is equivalent to defining the number of steps for which HyQ requires to reach to the goal position. To this end, we assume a virtual average stride length and based on the goal position displacement, we calculate the number of steps. In this experiment, we choose average stride length of $35$~[cm]. We have tried different average stride lengths on simulation and we ultimately choose $35$~[cm]. However in general with the higher values, we can observe more dynamic motions. 

\begin{figure}[t]
	\includegraphics[width=1\columnwidth]{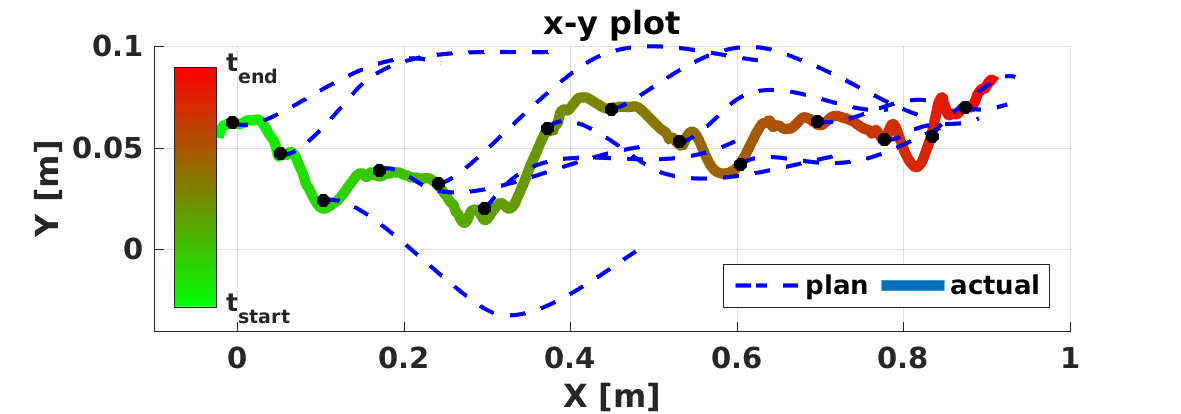}
	\vspace{-4mm}
	\caption{Overhead plot of continuously recomputed MPC paths for CoM (blue dashed line) and the executed path (green to red gradient) of a go-to task on HyQ. The time horizon is adapted online based on the goal distance. The $1$~[m] trotting forward command is received at $t_{start}$ and after s single update of the MPC optimizer (under $20$~[ms]) the plan for the next two steps are calculated and send it to robot. The computed plans are down-sampled and the plans are truncated to half of their total length.}
	\label{fig:xy_plot}
\end{figure}

Fig.~\ref{fig:xy_plot} shows the overhead view of this motion in $xy$ plane using a gradient color scheme which reflects the time evolution of the motion. A subsampled set of computed plans is also demonstrated with dashed blue lines. Here, we have only plotted the first half of the plans. This graph demonstrates that the realized CoM position and the plans are smooth and the MPC plans try to guide the robot toward the final goal position. 

\begin{figure}[t]
	\includegraphics[width=1\columnwidth]{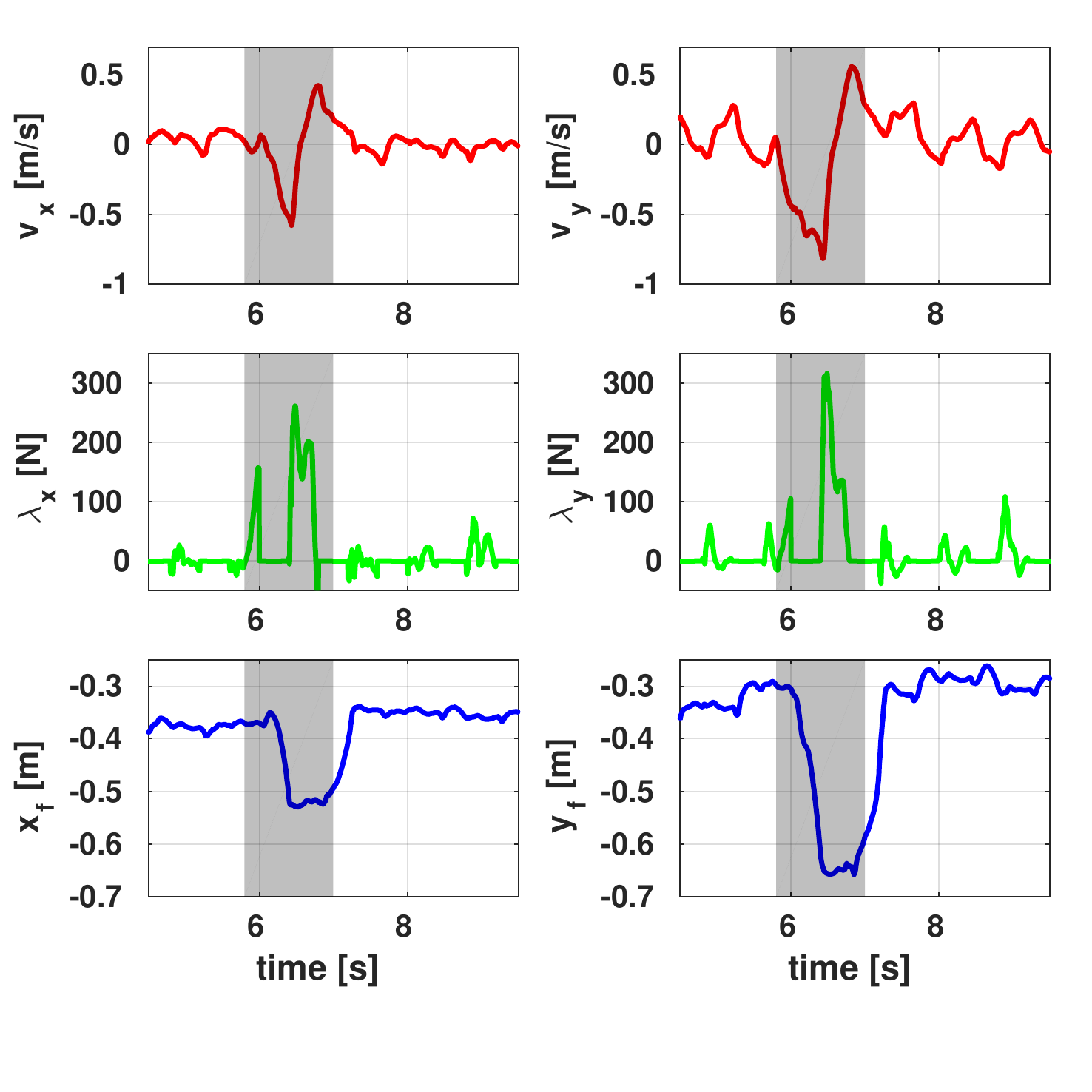}
	\vspace{-15mm}
	\caption{CoM velocity and the left-hind leg's contact force and position in the $x$ and $y$ directions. The gray area shows the trajectories at the time around a disturbance, where the disturbance force results is a sudden increase in the CoM velocity. The MPC plan for the contact forces and the foothold position nicely tries to stabilize the CoM.}
	\label{fig:side_pull}
\end{figure}

\begin{figure*}[t]
	\includegraphics[width=\textwidth]{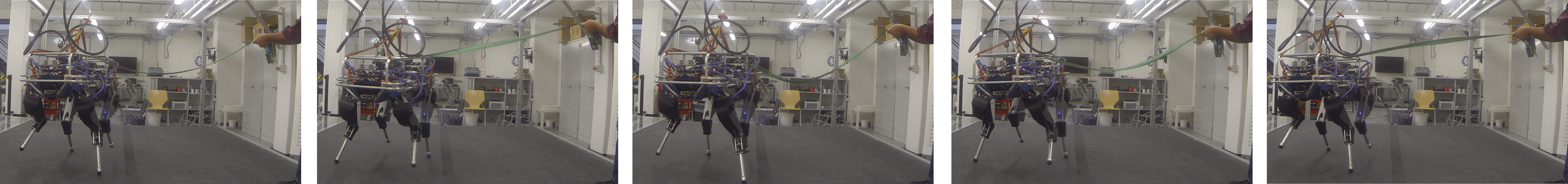}
	\vspace{-4mm}
	\caption{Disturbance rejection task. We insert a lateral disturbance by strongly pulling the robot from its right during trotting (second snapshot). In order to keep its balance, HyQ demonstrates a side-stepping motion in the third snapshot. Then, it eventually moves back to its initial position. The side-stepping in this experiment is about 40~cm.
	}
	\label{fig:time_lapse}
\end{figure*}

\paragraph{Disturbance rejection task} In this task, we command HyQ to trot in-place. For evaluating the disturbance rejection capability of our MPC, We pull/push the robot sidewise, i.e. $y$ direction (refer to video). Fig.~\ref{fig:side_pull} shows the time around one of these unknown disturbances (the gray area), where the disturbance force results is a sudden increase of velocities (in particular the $y$ direction). This plot also shows the $xy$ contact forces as well as the foot $xy$ motion in the global frame for a specific leg (left-hind). The robot uses two different strategies to reject the unknown external force resulted from the pulling. During the stance phase, it manipulates the contact forces in a way to resist the disturbance force while respecting the friction cone. Then in the swing phase, it reacts to the external force through a side-stepping motion. Fig.~\ref{fig:time_lapse} shows the snapshots of this experiment. In the video of this task, you can see that the robot maintains its balance against relatively strong pulls. The video also demonstrates the performance of the robot in response to the disturbances in the $x$ direction.

\subsection{\textsc{FastSLQ} Benchmarking}
Table~\ref{tab:performance} shows the frequency of the MPC loop for a various number of threads and number of subsystems (number of phases ahead) for planning. By looking at each column of this table, we notice that as the number of subsystems doubles the frequency reduces almost to half. This is due to the linear computational complexity of SLQ with respect to the optimization time horizon. As discussed in subsection~\ref{sec:disjointed_slq}, an important characteristic of \textsc{FastSLQ} is that it scales more favorably with respect to the time horizon. By comparing the values with the same color in Table~\ref{tab:performance} this characteristic becomes evident. We can see that while the time horizon is doubled the frequency does not reduce to half since the \textsc{FastSLQ} algorithm nicely benefits from the extra computational power (extra threads). 

\begin{table}[t]
\centering 
\caption{Comparison between the frequency of the MPC algorithm for different number os subsystems (number of partitions) and different number of threads. The value are in Hz.}
\label{tab:performance}
\begin{tabular}{c c  c  c  c  }
 Min Num. Subsystems & & Num. Threads &   \\
 \hline
  				& 1            & 2            & 4  \\
  2  			& {\color{blue}$31.9\pm0.7$} & {\color{red}$40.4\pm0.3$} & $61.7\pm1.6$  \\

  4  			& $17.5\pm0.2$ & {\color{blue}$25.2\pm0.3$} & {\color{red}$32.3\pm1.4$}  \\
  \hline
\end{tabular}
\end{table}

\begin{figure}[t]
	\includegraphics[width=1.05\columnwidth]{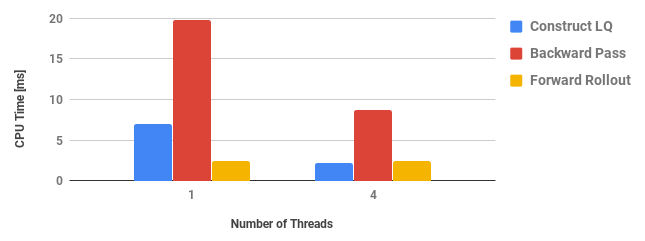}
	\vspace{-4mm}
	\caption{Comparison of the required CPU time for three main operations of SLQ with 1 thread and \textsc{FastSLQ} with 4 threads. The values are calculated by averaging over 1500 iterations of the algorithms in the MPC loop.}
	\label{fig:performance}
\end{figure}

To better understand this, Fig.~\ref{fig:performance} demonstrates the average CPU time required for the three main operations of SLQ and \textsc{FastSLQ}. 
As you see on the left, the computational bottleneck of SLQ is its backward pass which is almost $65\%$ of the total computation. \textsc{FastSLQ}, in contrast, leverages fully from the extra processing threads and drastically reduces the computation load of the backward pass (Fig.~\ref{fig:performance} right graph).

\section{Conclusion}
In this paper, we have introduced a real-time, constrained, nonlinear MPC approach for the motion planning of legged robots. The proposed approach uses a constrained SLQ algorithm in order to solve the MPC optimization problems. Moreover, we have introduced the \textsc{FastSLQ} algorithm which allows us to calculate the backward pass of SLQ algorithm in parallel. This drastically reduces the computational complexity of the backward pass which in turn improves the MPC loop frequency.   

The \textsc{FastSLQ-MPC} algorithm introduced in this paper can generate optimized trajectories for the next few phases of the motion within only a few milliseconds. This work shows the first application of whole-body MPC on legged robots for generating periodic gait patterns. We demonstrate that \textsc{FastSLQ-MPC} can be run at rates that exceed the state of the art by an order of magnitude. For example, in the case of 2 subsystems, the MPC loop can run at about $60$~Hz. The performance of our \textsc{FastSLQ-MPC} motion planner has been tested on both hardware and simulation for generating trotting gait. The capability of the planner for tracking user-defined goal as well as disturbance rejection is nicely shown on hardware and simulation. 


\section*{Acknowledgment} \footnotesize{This research has been supported in part by a Max-Planck ETH Center for Learning Systems Ph.D. fellowship to Farbod Farshidian and a Swiss National Science Foundation Professorship Award to Jonas Buchli and the NCCR Robotics.}

\bibliographystyle{bibliography/IEEEtran} \bibliography{bibliography/references}

\end{document}